\documentclass[letterpaper, 10 pt, conference]{ieeeconf}  

\IEEEoverridecommandlockouts                              %
\usepackage{graphicx} 
\usepackage{epsfig} 
\usepackage{soul,color} 
\usepackage{xcolor}
\usepackage{comment}
\usepackage{xspace}
\usepackage{mathtools}
\usepackage{booktabs}
\usepackage{multirow}
\usepackage{enumerate}
\usepackage{balance}
\usepackage{gensymb}
\usepackage{tabulary}
\usepackage{dblfloatfix}
\usepackage{layouts}
\usepackage{algorithm}
\usepackage{algpseudocode} 
\usepackage{cite}
\usepackage{bm}
\usepackage{hyperref}
\usepackage{svg}
\usepackage{cite}
\usepackage{caption}
\usepackage{subcaption}
\usepackage{colortbl}
\usepackage{overpic}
\usepackage{rotating}
\usepackage{afterpage}
\usepackage{float}
\usepackage{multirow}
\usepackage{url}
\usepackage{wrapfig}
\usepackage{listings}
\usepackage{adjustbox}
\usepackage{mdwtab}
\usepackage{array}
\usepackage{setspace}
\usepackage{mathdefs}
\usepackage{sonar}
\usepackage{units}

\usepackage[many]{tcolorbox}

\newcommand{\spatialpoint}{\ensuremath{\mathbf{x}}}
\newcommand{\arcpointatp}{\ensuremath{\mathbf{P}_p}}

\def\baselinestretch{0.955}\selectfont

\newcolumntype{?}{!{\vrule width 2pt}}
\definecolor{lightgreen}{rgb}{0.8,1,0.8}

\usepackage{caption}
\graphicspath{{../figs/}}

\makeatletter
\newcount\my@repeat@count
\newcommand{\myrepeat}[2]{%
  \begingroup
  \my@repeat@count=\z@
  \@whilenum\my@repeat@count<#1\do{#2\advance\my@repeat@count\@ne}%
  \endgroup
}
\title{\LARGE \bf
Neural Implicit Surface Reconstruction using Imaging Sonar
}

\author{Mohamad Qadri, Michael Kaess, and Ioannis Gkioulekas 
\thanks{$^{1}$M. Qadri, M. Kaess, and I. Gkioulekas are with The Robotics Institute, Carnegie Mellon University, USA.
        {\tt\small \{mqadri, kaess, igkioule\}@cs.cmu.edu}}%
\thanks{This work was partially supported by the Office of Naval Research grant N00014-21-1-2482 and National Science Foundation award 1730147. Ioannis Gkioulekas was supported by a Sloan Research Fellowship.}%
\thanks{$^{2}$The code will be made available at \url{https://github.com/rpl-cmu/neusis} }
}

\begin{document}

\maketitle
\thispagestyle{empty}
\pagestyle{empty}


\begin{abstract}
We present a technique for dense 3D reconstruction of objects using an imaging sonar, also known as forward-looking sonar (FLS). Compared to previous methods that model the scene geometry as point clouds or volumetric grids, we represent the geometry as a neural implicit function. Additionally, given such a representation, we use a  differentiable volumetric renderer that models the propagation of acoustic waves to synthesize imaging sonar measurements. We perform experiments on real and synthetic datasets and show that our algorithm reconstructs high-fidelity surface geometry from multi-view FLS images at much higher quality than was possible with previous techniques and without suffering from their associated memory overhead.
\end{abstract}

\section{Introduction}
\noindent
Imaging or forward-looking sonar (FLS) is an extensively
used sensor modality by Autonomous Underwater Vehicles
(AUV). The key motivation for using FLS sensors is their ability to provide long-range measurements, unlike optical cameras whose range is severely limited in turbid water---a common situation in the field. Their versatility has resulted in their incorporation as a core sensor modality in applications including robotic path planning~\cite{ho2018virtual,sodhi2019online}, localization~\cite{johannsson2010imaging, negahdaripour20133, westman2018feature, yang2017acoustic,arnold2018robust}, and the automation of tasks potentially dangerous or mundane for humans such as underwater inspection~\cite{albiez2015flatfish} and mapping~\cite{aykin2016three, huang2016incremental, negahdaripour2018application, wang2019underwater}.  

FLS outputs 2D image measurements of 3D structures by emitting acoustic pulses and measuring the energy intensity of the reflected waves. While the sonar resolves azimuth and range, the elevation angle is ambiguous, and an object at a specific range and azimuth can be located anywhere along the elevation arc. Hence, the task of 3D reconstruction using FLS measurements can be equivalently framed as the task of resolving the elevation ambiguity from the image readings. 

Existing algorithms for 3D reconstruction from FLS measurements can be grouped into geometry-based, physics-
based and, more recently, learning-based methods. How-
ever, most existing approaches either place restrictions on
the robotic/sensor setup (elevation aperture, motion of the
vehicle, etc.); rely on volumetric grids that are prohibitively
expensive for large scenes or scenes with fine-grained geometry; or, specific to learning approaches, require the use
of large labeled training sets that are difficult to collect in
underwater environments.

To address these shortcomings, we approach the problem of underwater FLS-based 3D reconstruction through the lens of differentiable rendering and leverage the representational power of neural networks to encode the imaged object as an implicit surface. Our overall reconstruction
approach comprises the following components:
\begin{itemize}
    \item A differentiable volumetric renderer that models the propagation of acoustic spherical wavefronts.
    \item A representation of 3D surfaces as zero-level sets of neural implicit functions.
    \item  A regularized rendering loss for 3D reconstruction using
imaging sonars.
\end{itemize}
To the best of our knowledge, this work is the first to
introduce a physics-based volumetric renderer suitable for
dense 3D acoustic reconstruction using wide-aperture imag-
ing sonars. We evaluate our approach against different unsupervised methods on
simulated and real-world datasets, and show that it outperforms
previous state of the art. We will open-source our code together with different datasets$^2$.
\section{Related Work}
\subsection{3D Reconstruction Using Imaging Sonar}  
\noindent
Different methods have been introduced to produce both sparse \cite{mai20173, huang2015towards, huang2016incremental, negahdaripour2018application, westman2019degeneracy, westman2018feature} and dense 3D reconstructions using FLS. The focus of this work is on dense object-level 3D reconstruction. A number of algorithms enforced assumptions or constraints on the physical system to obtain reliable 3D models. Teixeira et al.~\cite{teixeira2016underwater} successfully reconstructed a 3D map of a ship hull by leveraging probabilistic volumetric techniques to create submaps which are later aligned using Iterative Closest Point (ICP). However, the sonar aperture was set to $1^{\degree}$ and all detected points were assumed to lie on the zero-elevation plane which leads to reconstruction errors and prohibits extending the method to larger apertures. A line of work  \cite{joe2022probabilistic, joe2021sensor, mcconnell2020fusing, mcconnell2021predictive}  uses two complementary sensors (imaging and profiling sonars) and performs sensor fusion to disambiguate the elevation angle. In our work, we restrict our setup to a single imaging sonar. Westman et al.~\cite{westman2020theory} proposed a method to reconstruct specific points on surfaces (aka. Fermat Paths). However, it places constraints on the vehicle's motion as it needs a view ray perpendicular to the surface at each surface point and hence, requires a large number of images collected from specific views.

Another set of methods uses generative models to obtain dense 3D reconstructions. Aykin et al.~\cite{aykin20153}, \cite{aykin2016modeling} attempt to estimate the elevation angle of each pixel by leveraging information from both object edges and shadows which restricts the object to be on the seafloor. Westman et al.~\cite{westman2019wide} further extended the idea to do away with the seafloor assumption but still required estimates of object edges. Negahdaripour et al.~\cite{negahdaripour2017refining} proposed an optimization-based algorithm to refine an initial 3D reconstruction obtained using space carving by encouraging consistency between the actual sonar images and the images produced by the generative model. However, generative methods generally rely on assumptions of the surface reflectivity proprieties and on 3D estimates of object edges which makes them impractical in real scenarios.

Various volumetric methods have also been proposed. Wang et al.~\cite{wang20183d} introduced an inverse sonar sensor model to update the occupancy in a voxel grid and later extended it to handle errors in pose estimates by aligning local submaps using graph optimization \cite{wang2019three}. Although these methods, as probabilistic frameworks, can be more robust compared to space carving techniques \cite{aykin20153, aykin2016three}, they consider each voxel independently and ignore inherent surface constraints. Guerneve et al.~\cite{guerneve2015underwater} frame the problem of 3D volumetric reconstruction as a blind deconvolution with a spatially varying kernel which can be reformulated as a constrained linear least squares objective. However, the method makes a linear approximation to the vertical aperture and places restrictions on the motion of the sonar limiting its practical application. Westman et al.~\cite{westman2020volumetric} noted the equivalence of 3D sonar reconstruction to the problem of Non-Line-of-Sight (NLOS) imaging. It introduced a regularized least square objective and solved it using the alternating direction method of multipliers (ADMM). All aforementioned volumetric methods, however, share similar limitations since extracting high-fidelity surfaces from volumetric grids is difficult. These approaches can also be computationally expensive for larger scenes or a fine discretization of volumes. 

More recently, learning-based methods were proposed to resolve the elevation ambiguity. DeBortoli et al.~\cite{debortoli2019elevatenet} proposed a self-supervised training procedure to fine-tune a Convolutional Neural Network (CNN) trained on simulated data with ground truth elevation information. Wang et al.~\cite{wang2021elevation} use deep networks to transfer the acoustic view to a pseudo frontal view which was shown to help with estimating the elevation angle. However, these methods are limited to simple geometries or require collecting a larger dataset of real elevation data. Arnold et al.~\cite{arnold2022spatial} propose  training a CNN to predict the signed distance and direction to the
nearest surface for each cell in a 3D grid. However, the method requires ground truth Truncated Signed Distance Field (TSDF) information which can be difficult to obtain.

In this work, we propose a physics-based renderer which uses raw FLS images and known sonar pose estimates to represent objects as zero-level sets of neural networks. It does not require hand-labeled data for training nor does it place restrictions on the setup or environment (voxel size, need for reflectance information, etc.)

\subsection{Neural Implicit Representation}
\noindent
NeRF \cite{mildenhall2021nerf} introduced a volume rendering method to learn a density field aimed at novel view synthesis. It samples points along optical rays and predicts an output color which is then checked against that of the ground truth pixel. IDR \cite{yariv2020multiview} introduced a surface rendering technique that contrary to the volume rendering technique of NeRF, only considers a single point intersection on a surface. Hence, it fails to properly capture areas of abrupt changes in the scene. NeuS \cite{wang2021neus} leveraged the volume rendering technique of NeRF to perform 3D surface reconstructions and showed impressive results against state-of-the-art neural scene representation methods for scenes with severe occlusions and complex structures. NeTF \cite{shen2021non} applied ideas from NeRF to the problem of NLOS imaging which was shown in \cite{westman2020volumetric} to have close similarity to FLS 3D reconstruction.
All these methods focus on 3D reconstruction using optical sensors, either intensity or time-of-flight based. Our focus is on learning surfaces from acoustic sonar images.
\section{Approach}

\subsection{Image Formation Model}
\begin{figure}[h!]
\begin{subfigure}[h]{0.18\textwidth}
\centering
\includegraphics[width=1\columnwidth, height=4.5cm]{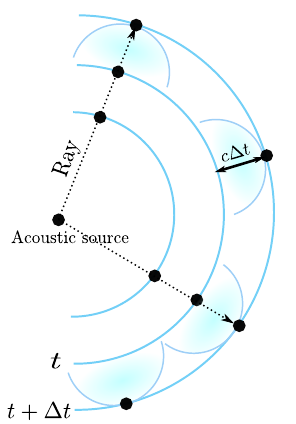}
\caption{}
\label{propagationModelA}
\end{subfigure}
\begin{subfigure}[h]{0.18\textwidth}
\centering
\vspace{2cm}
\includegraphics[width=1\columnwidth, height=2.5cm]{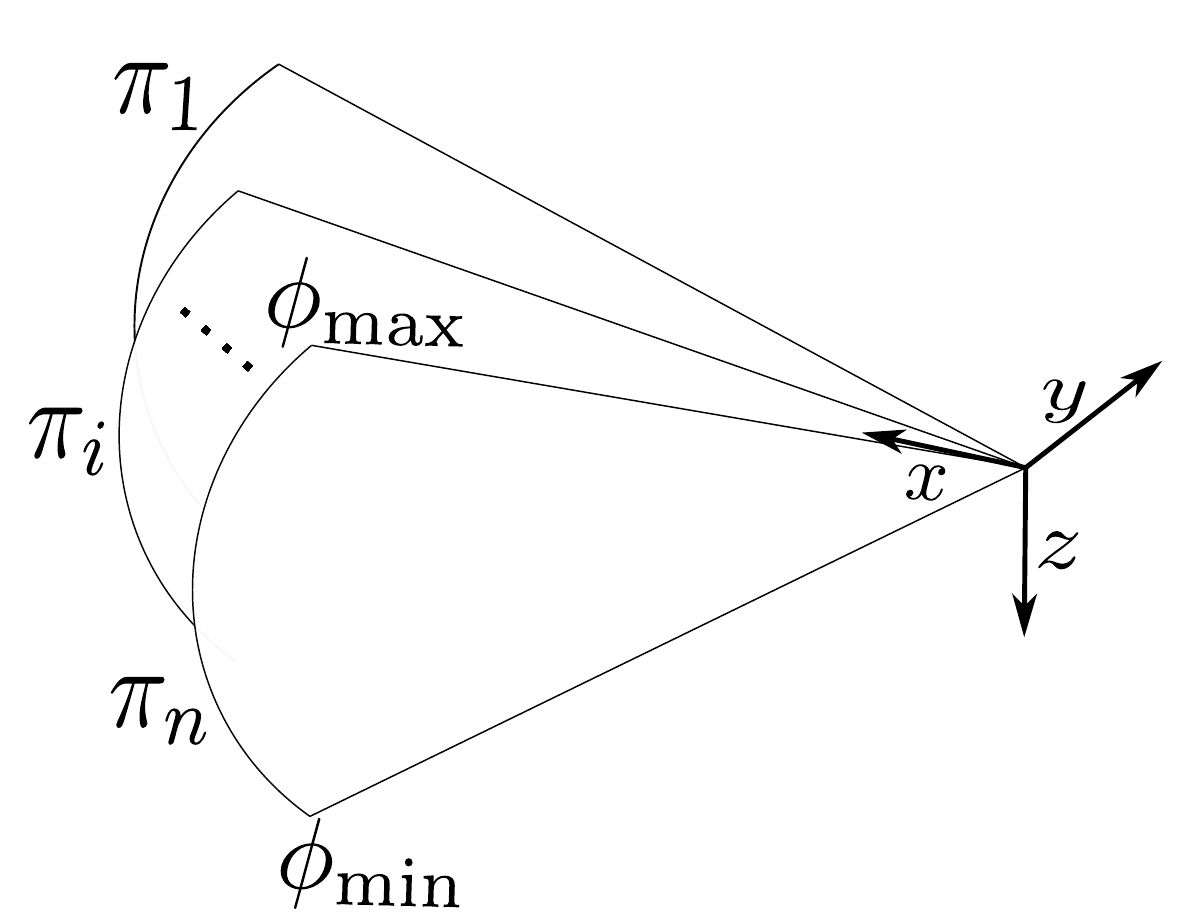}
\caption{}
\label{sonarPlanes}
\end{subfigure}
\begin{subfigure}[h]{0.09\textwidth}
\centering
\includegraphics[width=1\columnwidth, height=4.5cm]{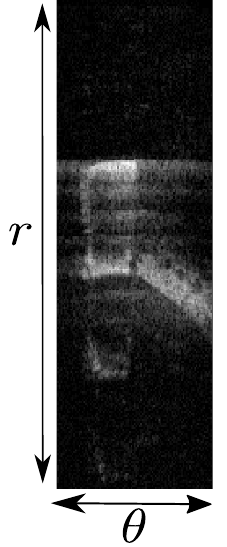}
\caption{}
\label{exampleImage}
\end{subfigure}%
\caption{(a) Sound propagates as spherical wavefronts. An acoustic ray is defined as the ray starting at the acoustic source and terminating at the wavefront (figure inspired by the \textit{Discovery of Sound in the Sea} project \cite{vigness2010discovery}) (b) Each image column $\theta_i$ is the projection of the circular arc $\pi_i$ onto the plane $z=0$. (c) Example of a sonar image. Each pixel at $(r, \theta)$ corresponds to the intensity reading of all points along the elevation arc.}
\label{propagationModel}
\end{figure}

An FLS 2D image $\mathcal{I}$ comprises pixels corresponding to discretized range and azimuth $(r_i, \theta_i)$ bins. Each pixel value is proportional to the sum of acoustic energy from all reflecting points $\{\mathbf{P}_i = (r_i, \theta_i, \phi_i); \phi_{\mathrm{min}} \leq \phi_i \leq \phi_{\mathrm{max}} \}$, $\phi_i$ being the elevation angle (Fig.~\ref{exampleImage}). However, the elevation angle $\phi_i$ is lost since each column $\theta_i$ of an FLS image is the projection onto the $z=0$ plane of a circular sector $\pi_i$ constrained to the sonar vertical aperture $(\phi_{\mathrm{min}}, \phi_{\mathrm{max}})$ and containing the $z$ axis (Fig.~\ref{sonarPlanes}). 

\noindent

\begin{figure}[h!]
\centering
  \includegraphics[width=1\columnwidth]{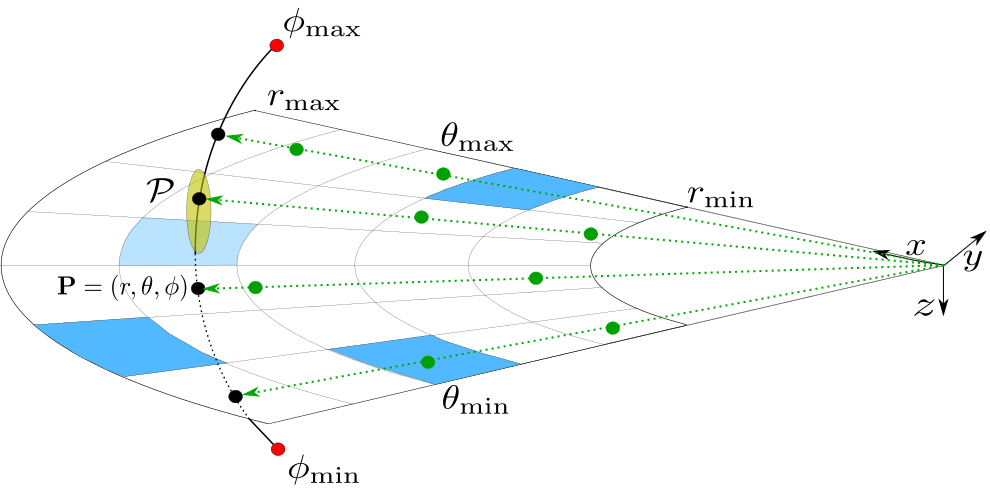}
  \caption{1) All points $\mathbf{P} = (r, \theta, \phi)$ on the arc are projected onto the $z=0$ elevation plane. 2) An example of an infinitesimally small patch on the arc $\mathcal{P}$  is shown in yellow. 3) Illustrating our sampling scheme: sampled pixels are colored in blue. Sampled points on the arc are shown in black. For each point on the arc, we construct the acoustic ray (green arrow) and sample points on each ray (green points).}
  \label{sonargeom}
 \end{figure}

We now present our rendering equation. Imaging sonars are active sensors that emit a pulse of sound and measure the strength of the reflected acoustic energy. Let $\textit{E}_e$ be the emitted acoustic energy by the sonar. Now, consider a unique infinitesimal reflecting patch $\mathcal{P}_i$ ``illuminated" by the acoustic wave and located on the arc $\mathcal{A}(\phi) \in \pi_i$ which passes through $(r_i, \theta_i, 0)$ (Fig.~\ref{sonargeom}). The reflected acoustic energy at $\mathcal{P}_i$ and received by the sonar can be approximated as:
\begin{align}
   & \textit{E}_r(r_i, \theta_i, \phi_i)\!=\!\int_{r_i-\epsilon}^{r_i+\epsilon} \frac{\textit{E}_e}{r^2}  \underbrace{e^{-\int_0^{r_i} \sigma(r', \theta_i, \phi_i ) \text{d}r'}}_{T} \sigma(r, \theta_i, \phi_i ) r\text{d}r
   \label{onepatchreflected}
\end{align}
where $2\epsilon$ is the patch thickness, $\sigma$ is the particle density at $\mathcal{P}_i$, and the factor $\frac{1}{r^2}$ accounts for spherical spreading on both the transmit and receive paths. $T$ is the transmittance, corresponding to exponential attenuation of a wave due to particle absorption---equivalently, the probability that the acoustic wave travels between two points unoccluded. We note that, when the sonar emitter and receiver are collocated, this probability is identical during the transmit (sonar to $\mathcal{P}_i$) and return ($\mathcal{P}_i$ to sonar) paths; thus, transmittance is accounted for only once for both paths. This is analogous to the effect of coherent backscattering in optical wave propagation with collocated emitter and receiver~\cite{mishchenko2006multiple}.



Now consider a surface composed of many such patches. The received energy by the sonar is simply the sum of the reflected energy by all patches $\{ \mathcal{P}_i\} \in \mathcal{A}(\phi)$ which approximate the surface. Hence, we  arrive at the following image formation model:
\begin{align}
& I(r_i, \theta_i) = \int_{\phi_\text{min}}^{\phi_\text{max}} \int_{r_i-\epsilon}^{r_i+\epsilon} \frac{\textit{E}_e}{r^2} e^{-\int_0^{r_i} \sigma(r', \theta_i, \phi ) \text{d}r'} \sigma(r, \theta_i, \phi) r\text{d}r\text{d}\phi \nonumber\\ 
& \quad \quad \quad  \;= \int_{\phi_\text{min}}^{\phi_\text{max}} \int_{r_i-\epsilon}^{r_i+\epsilon} \frac{\textit{E}_e}{r} T(r, \theta_i, \phi) \sigma(r, \theta_i, \phi) \text{d}r\text{d}\phi.
\label{renderingloss}
\end{align}

Note that although sound propagation through liquids is fundamentally different from that of light (longitudinal vs.~transverse waves), different geometric acoustic modeling techniques still borrowed heavily from graphics and ray optics \cite{siltanen2007room}. These methods fundamentally rely on Huygen's  principle of  sound travel through mediums which approximates the spherical wavefront as many energy-carrying particles travelling at the speed of sound. Hence, analogous to the concept of a light ray, we view an acoustic ray as the ray starting at the sonar acoustic center and ending at $\mathcal{P}_i$ (Fig.~\ref{propagationModelA}).
\subsection{Rendering Procedure}
\noindent
Similarly to Yariv et al.~\cite{yariv2020multiview} and Wang et al.~\cite{wang2021neus}, we represent the sonar-imaged surface using two multi-layer perceptrons (MLPs): a neural implicit surface, $\mathbf{N}$, which maps a spatial coordinate $\spatialpoint =(r, \theta, \phi)$ to its signed distance; and a neural renderer, $\mathbf{M}$, which outputs the outgoing radiance at $\spatialpoint$.

Once the surface $\mathcal{S}$ is learned, we can extract it as the zero level set of $\mathbf{N}$:
\begin{align}
    \mathcal{S} = \{\spatialpoint\in\mathbb{R}^3: \mathbf{N}(\spatialpoint) = 0 \}.
\end{align}
To train our network using the rendering loss (Eq. \ref{renderingloss}), we leverage the following equation from Wang et al.~\cite{wang2021neus} to estimate the value of the density $\sigma(\spatialpoint)$ from the SDF:
\begin{align}
    \sigma(\spatialpoint) = \max \left( \frac{\frac{ -\text{d} \Phi_s (\mathbf{N}(\spatialpoint)) }{\text{d}\spatialpoint}}{\Phi_s (\mathbf{N}(\spatialpoint)) }, 0 \right)
\end{align}
where $\Phi_s(\tau) \equiv (1 + e^{-s\tau})^{-1}$ is the sigmoid function used as a smooth approximator of the occupancy indicator function $\mathcal{O}(\spatialpoint) \equiv \mathbf{1}[\mathbf{N}(\spatialpoint) \geq 0] $. 

\subsection{Sampling Procedure}
\noindent 
Existing work that targets optical cameras leverages ray optics where sampling points along a ray originating at some pixel is sufficient to approximate the rendering loss. On the contrary, our rendering loss in Eq.~\ref{renderingloss} requires producing point samples along the arc at $p_i=(r_i, \theta_i)$ as well as  samples along each acoustic ray. To obtain a balanced dataset of zero and non-zero intensity samples when processing an image, we sample $\mathbf{N}_{\mathcal{P}^1}$ random image pixels as well as  $\mathbf{N}_{\mathcal{P}^2}$ pixels with an intensity $I(r_i, \theta_i)$ greater than a threshold. Let $\mathcal{P}$ be the set of sampled pixels.

For each pixel $p_i \in \mathcal{P}$, we use stratified sampling to obtain samples along the arc. 
We discretize the elevation range  $[-\phi_{\text{min}}, \phi_{\text{max}}]$ into $\mathbf{N}_{\mathcal{A}}$ equally spaced angles. Hence, the difference between two consecutive angles is $\Delta\phi = \frac{\phi_{\text{max}} - \phi_{\text{min}}}{\mathbf{N}_{\mathcal{A}}}$. We perturb these angles by adding $\mathbf{N}_{\mathcal{A}}$ randomly generated noise values $ \sim \text{Uniform}(0, 1)  \Delta\phi$ to obtain a set of points $\mathbf{A}_p = \{ \arcpointatp=(r_i, \theta_i, \phi_{\arcpointatp})) \}$ on the arc. 

\begin{figure}[h!]
\centering
  \includegraphics[width=1\columnwidth]{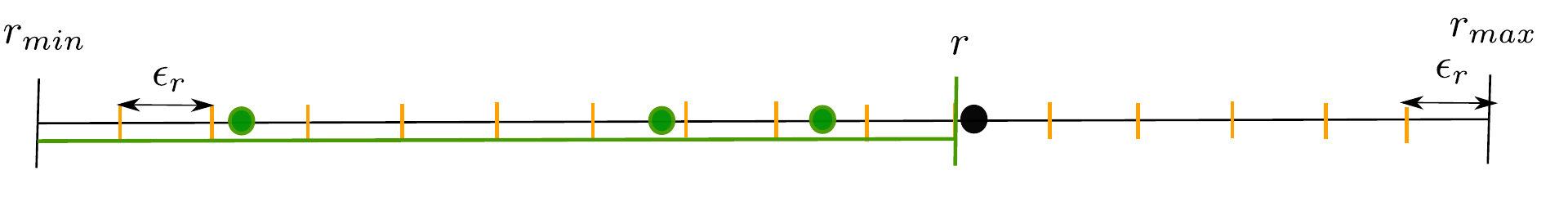}
  \caption{Sampling along radius $r$. We first sample range bins and 
  then sample one point in each bin (green points). This is the set $\mathbf{R}_{\arcpointatp}$. The black point is the perturbed point on the arc $\arcpointatp$.}
  \label{rsample}
  \vspace{-2mm}
 \end{figure}

For each sampled point $\arcpointatp$, we first construct the acoustic ray $\mathcal{R}_{\arcpointatp} $ which starts at the acoustic center of the sonar and terminates at $\arcpointatp$ and then sample $\mathbf{N}_{\mathcal{R}} - 1$ points along each ray. Specifically, we first sample $\mathbf{N}_{\mathcal{R}} - 1$ range values $r'$ such that $r' < r$ and $r'=i\epsilon_r $ for some $i>0$  ($\epsilon_r$ being the sonar range resolution). We obtain the set of points $\mathbf{R}_{\arcpointatp} = \{ \mathbf{p} = (r', \theta, \phi_{\arcpointatp}) \}$. The points $\mathbf{R}_{\arcpointatp} \cup \mathbf{A}_p$ constitute a set of $\mathbf{N}_{\mathcal{R}}$ points along the ray ($\mathbf{N}_{\mathcal{R}} - 1$ points along the ray + exactly 1 point on the arc). Finally, we perturb the range value of all points by adding uniformly distributed noise  $ \sim \text{Uniform}(0, 1)  \epsilon_r$ (Fig. \ref{rsample}).

Note that the points $\mathbf{R}_{\arcpointatp} \cup \mathbf{A}_p$ are expressed in spherical coordinates in the local sonar coordinate frame and hence need to be re-expressed in a global reference frame common to all sonar poses. We first transform the points to Cartesian coordinates:
 \begin{align}
   & x = r \cos(\theta) \cos(\phi) \\   
   & y = r \sin(\theta) \cos(\phi) \\
   & z = r \sin(\phi)
 \end{align}
 and then transform to world frame by multiplying with the sonar to world transform $T_W^{\text{sonar}} = \begin{bmatrix}
    R_W^{\text{sonar}} &  t_W^\text{{sonar}} \\
    \mathbf{0}^T & 1
 
 \end{bmatrix}$.
  
The resulting set of points expressed in world frame $\mathbf{R}^W_{\arcpointatp} \cup \mathbf{A}^W_p$ are used as inputs to the SDF neural network $\mathbf{N}$. Finally, the direction of each ray is defined by the unit vector 
\begin{align}
  D(\arcpointatp) = \frac{T_W^{\text{sonar}}\arcpointatp -  t_W^{\text{sonar}} }{|T_W^{\text{sonar}}\arcpointatp -  t_W^{\text{sonar}}|}  
\end{align}
 Let $\mathbf{X}$ be the set of all sampled points across all pixels, arcs and rays. This is the input batch to the neural network.

 \subsection{Discretized Image Formation Model}
\noindent
The discrete counterpart of the image formation model in Eq. \ref{renderingloss} is: 
\begin{align}
    \hat{I}(r, \theta) = \sum_{\arcpointatp \in \mathcal{A}_p} \frac{1}{r_{\arcpointatp}} T[\arcpointatp] \alpha[\arcpointatp] \mathbf{M}(\arcpointatp),
\end{align}
\noindent
where:
$\mathcal{A}_p$ is the arc located at $(r, \theta)$, $r_{\arcpointatp}$ is the range of the disturbed point $\arcpointatp$ on the arc, $\mathbf{M}(\arcpointatp)$ is the predicted intensity at $\arcpointatp$ by the neural renderer,
\begin{align}
\alpha[\mathbf{p}_i] = 1 - \exp\left(
- \int_{\mathbf{p}_i}^{\mathbf{p}_{i+1}} \sigma(p)\text{d}p
\right)
\end{align}
is the discrete opacity  \cite{wang2021neus} at a point $\mathbf{p}_i$ ($\mathbf{p}_i$ and $\mathbf{p}_{i+1}$ being consecutive samples along the ray) which was further shown to equal:
 \begin{align}
    \alpha[\mathbf{p}_i] = \max(\frac{\mathbf{\Phi}_s(\mathbf{N}(\mathbf{p}_{i}))-\mathbf{\Phi}_s(\mathbf{N}(\mathbf{p}_{i+1}))}{\mathbf{\Phi}_s(\mathbf{N}(\mathbf{p}_{i}))}, 0).
\end{align} 
Finally,  
\begin{align}
    T[\arcpointatp] = \prod_{\mathbf{p}^1 \in \mathbf{R}_{\arcpointatp}} (1-  \alpha[\mathbf{p}^1])
\end{align}
is the discrete transmittance value at $\arcpointatp$ (the endpoint of the ray). This is the product of one minus the opacity values $\alpha$  of all points on the acoustic ray excluding the  $\alpha$ at $\arcpointatp$.

\subsection{Training Loss}
\noindent
Our loss function is constituted of three terms: the intensity loss in addition to eikonal and $\ell_1$ regularization terms.
The intensity loss  
\begin{align}
\mathcal{L}_{\text{int}} \equiv \frac{1}{\mathbf{N}_\mathcal{P}^1 + \mathbf{N}_\mathcal{P}^2} \sum_{p \in \mathcal{P}}||\hat{I}(p) - I(p)||_1,
\end{align}
encourages the predicted intensity to match the intensity of the raw input sonar images. The eikonal loss \cite{gropp2020implicit}
\begin{align}
    \mathcal{L}_{\text{eik}} \equiv \frac{1}{\mathbf{N}_\mathcal{R}\mathbf{N}_\mathcal{A} (\mathbf{N}_\mathcal{P}^1 + \mathbf{N}_\mathcal{P}^2)} \sum_{\spatialpoint\in \mathbf{X} } (||\nabla \mathbf{N}(\spatialpoint)||_2 - 1)^2,    
\end{align}
is an implicit geometric regularization term used to regularize the SDF encouraging the network to produce smooth reconstructions. Finally, we draw inspiration from the NLOS volumetric albedo literature \cite{heide2014diffuse, westman2020volumetric}, and add the  $\ell_1$ loss term
\begin{align}
    \mathcal{L}_{\text{reg}} \equiv \frac{1}{\mathbf{N}_\mathcal{R}\mathbf{N}_\mathcal{A} (\mathbf{N}_\mathcal{P}^1 + \mathbf{N}_\mathcal{P}^2)} \sum_{\spatialpoint \in \mathbf{X}} || \alpha[\spatialpoint]||_1,
\end{align}
to help produce favorable 3D reconstructions when we use sonar images from a limited set of view directions. Hence, our final training loss term is:
\begin{align}
\mathcal{L} = \mathcal{L}_{\text{int}} + \lambda_{\text{eik}} \mathcal{L}_{\text{eik}} + \lambda_{\text{reg}} \mathcal{L}_{\text{reg}}.
\end{align}




\section{Network Architecture}
\begin{figure}[h!]
\centering
  \includegraphics[width=1\columnwidth]{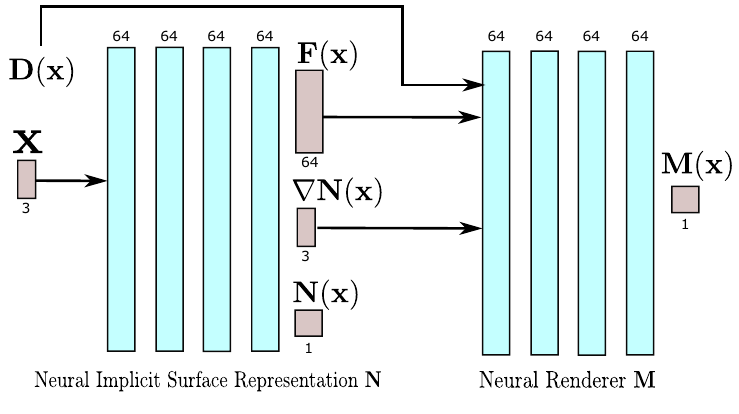}
  \caption{Our neural network architecture. The neural implicit representation $\mathbf{N}$ takes 3D spatial coordinates $\textbf{x}$ as input and outputs their signed distance to the surface as well as a learned feature vector $\mathbf{F}(\spatialpoint)$. We use PyTorch's autodiff \cite{paszke2017automatic} to compute $\nabla \mathbf{N}(\spatialpoint) $, the gradient of the signed distance at $\spatialpoint$.} 
  \label{architecture}
\end{figure}
\noindent 
We model $\mathbf{N}$ and $\mathbf{M}$ as two MLPs each with 4 hidden layers of size 64 (Fig \ref{architecture}). We additionally apply positional encoding to the input spatial coordinates and use weight normalization similar to IDR. While existing works that use optical cameras typically rely on larger networks to successfully learn high-frequency color and texture information, we found the proposed architecture to have sufficient capacity to learn different shapes from FLS images. Decreasing the size of the network was especially important to handle GPU memory overhead during training caused by the added sampling dimension (arcs). 

\section{Evaluation}
\noindent
As our comparison metric, we use the mean and root mean square (RMS) Hausdorff distance defined as:
\begin{align}
\begin{split}
    d_H(\mathcal{M}_1, \mathcal{M}_2) = \max ( \max_{\mathbf{p} \in \mathcal{M}_1} \min_{\mathbf{q} \in \mathcal{M}_2} || p - q ||_2 , \\ 
    \max_{\mathbf{q} \in \mathcal{M}_2} \min_{\mathbf{p} \in \mathcal{M}_1} || p - q ||_2  )
\end{split}
\end{align}
$\mathcal{M}_1 $ and $\mathcal{M}_2$ being respectively the ground truth (GT) and reconstructed meshes. We evaluate our method against Back-Projection (BP) and ADMM, two state-of-the-art optimization-based methods for unsupervised object-centric 3D reconstruction using imaging sonar. \footnote{We use the implementation of Westman et al.~\cite{westman2020volumetric}.}  BP is similar to the occupancy grid mapping method (OGM) as it uses the inverse sensor model to update the voxel occupancy while, however, ignoring the correlation between grid cells. We note that both ADMM and BP  generate a density field $\mathbf{F}(\sigma)$. Hence, for each possible density $\sigma$ (i.e., $\sigma \in [0, 1]$), we extract a surface using Marching Cubes and report the metrics based on the best $\sigma$ value. The mesh quality generated by ADMM also depends on the regularization weight terms which we empirically tuned for each object. With our approach, extracting the zero-level set of $\mathbf{N}(x)$ directly generates a high-quality mesh. However, for the purpose of metric generation, we also try different level-sets near zero: $\mathcal{S} = \{\spatialpoint \!\in\mathbb{R}^3\!: \mathbf{N}(\spatialpoint) = s \ | \ s  \in [-0.1, 0.1] \}$.
We run our experiments on a system with an NVIDIA RTX 3090 GPU, an Intel Core i9-10900K, and 32GB of RAM. Our network training time until convergence is $\sim6$ hours.

\subsection{Simulation}
\begin{figure} [h!]
 \centering
 \includegraphics[width=1\columnwidth]{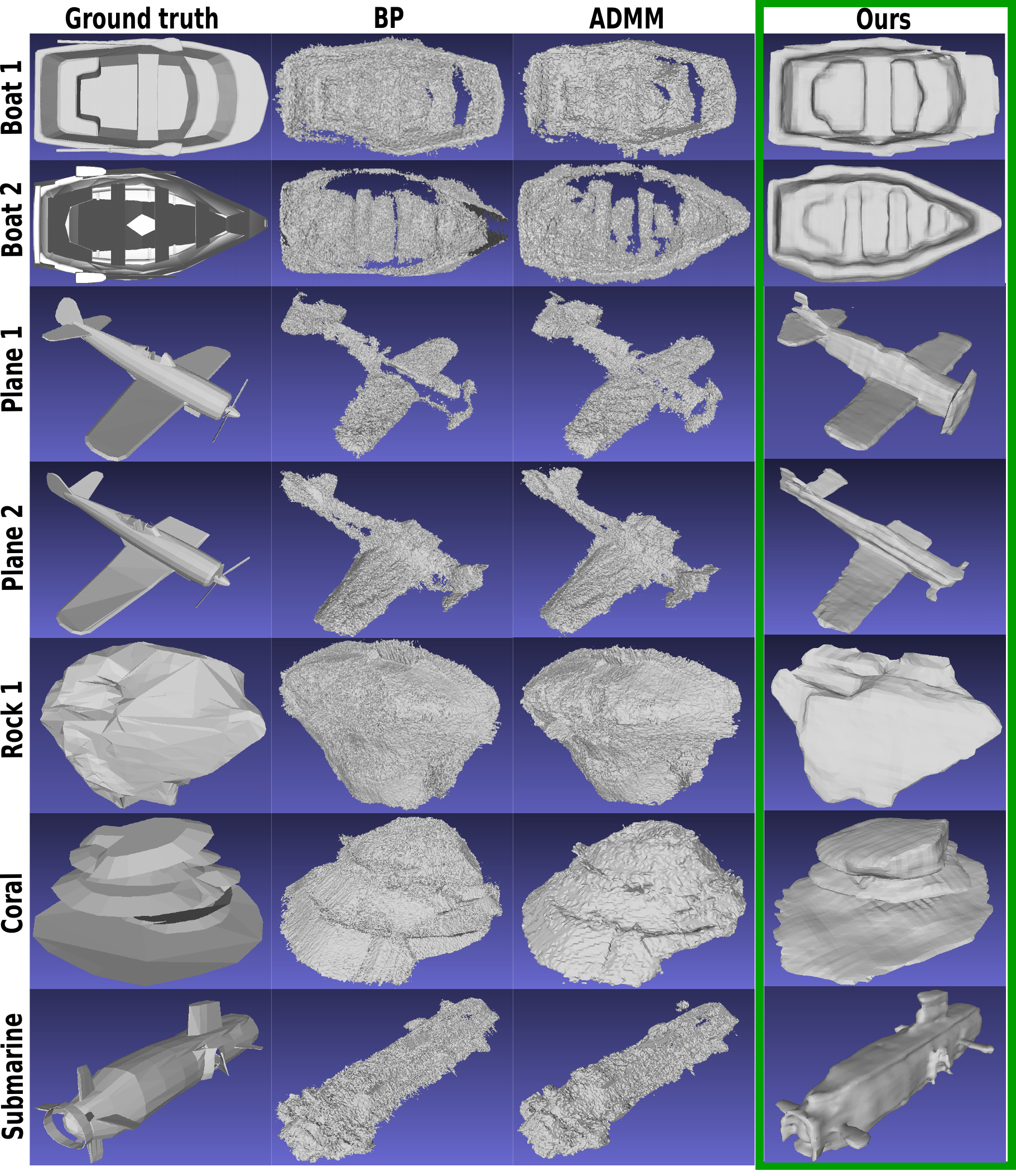}
\caption{3D reconstructions generated by each method using  simulated data from HoloOcean with a $14\degree$ elevation angle. Qualitatively, our method outputs more faithful 3D reconstructions compared to ADMM and BP.}
\label{fig:simQualitative}
\end{figure}

\noindent
We use HoloOcean \cite{potokar2022holoocean, Potokar22iros}, an underwater simulator to collect datasets of different objects of various shapes and sizes. We use the simulator's default noise parameters; namely a multiplicative noise $w^{\text{sm}} \sim \mathcal{N}(0, 0.15)$ and additive noise $w^{\text{sa}} \sim \mathcal{R}(0.2)$ (where $\mathcal{R}$ is the Rayleigh distribution and parameters are in units of normalized pixel intensity in the range $[0, 1]$). We also enable the simulation of multipath effects. The maximum range of the sonar was set to 8m. Before feeding the raw data to the three algorithms, we perform minimal filtering of speckle noise in the images by zeroing out pixels whose intensities are less than a threshold.  After generating the meshes, we align them to the GT using ICP and  report in Table \ref{table:simtable} the mean and RMS Hausdorff distance to the GT for different objects and two sonar vertical apertures ($14\degree$ and $28\degree$). Figure~\ref{fig:simQualitative} shows example 3D reconstructions obtained using each algorithm.

We see that our method produces more accurate reconstructions compared to the baselines in terms of  3D reconstruction accuracy and mesh coverage. 
The neural network implicit regularization combined with the eikonal loss favors learning smooth surfaces while avoiding bad local minimas when the input images potentially do not contain enough information to completely constrain and resolve the elevation of every 3D point in space.

For large objects (specified by an asterisk in the table), we decreased the grid voxel resolution of the baseline methods by one-half (increased the voxel size from the default value of 0.025m to 0.05m) to prevent the system from running out of memory (OOM): the ADMM and BP baselines do not leverage stochastic updates and hence, need to construct the optimization objective by processing all images in one go. This leads to memory overhead for larger objects, objects that require a fine discretization of the volume, or in the presence of a large number of non-zero pixel intensities \footnote{A re-implementation of the baselines which solves the optimization problem using stochastic updates can help dealing with OOM errors.}. In contrast, we train our renderer on a different subset of images in every iteration and use stochastic updates (the Adam optimizer) to optimize the function which significantly reduces memory requirements.

\begin{table}[h!]
\vspace{2mm}
    \begin{center}
    \resizebox{1\linewidth}{!}{
        \begin{tabular}{c|c|c|c?c|c?c|c}
            \toprule
            \multicolumn{2}{c|}{} &  \multicolumn{2}{c|}{BP}&  \multicolumn{2}{c|}{ADMM}&  \multicolumn{2}{c|}{Ours}  \\
            \hline
            \multicolumn{2}{c|}{} &  RMS & Mean &  RMS & Mean &  RMS & Mean  \\
            \hline
    Boat 1 & $14\degree$ & 0.092 &  0.068 &  0.100 & 0.073 & \textbf{0.055} & \textbf{0.042}\\
    $(3.8 \times 1.7\times 0.84)$ &$28\degree$ &0.196& 0.149 & 0.136 & 0.101 & \textbf{0.063} & \textbf{0.046} \\
        \hline
    Boat 2& $14\degree$ & 0.121 & 0.090 & 0.084 & 0.064 & \textbf{0.076} & \textbf{0.062}\\
    $(5.7 \times 2.3\times 1.2)$ &$28\degree$ &0.101 & 0.071 & 0.111 & 0.081 & \textbf{0.083} & \textbf{0.068} \\
    \hline
    Plane 1& $14\degree$ & $0.204^*$ & $0.138^*$ & $0.191^*$ & $ 0.147^*$ & \textbf{0.160} & \textbf{0.096}\\
    $(13.5 \times 11.5\times 3.6)$&$28\degree$ &$0.256^*$& $0.206^*$ & $ 0.236^*$ & $  0.165^*$ & \textbf{0.167} & \textbf{0.098} \\
        \hline
    Plane 2& $14\degree$ & $0.204^*$ & $0.167^*$ & $0.181^*$ & $0.139^*$ & \textbf{0.122} & \textbf{0.082}\\
    $(9.1 \times 12.6\times 3.0)$&$28\degree$ &$0.333^*$& $0.251^*$ & $0.313^*$ & $0.224^*$ & \textbf{0.166} & \textbf{0.116} \\        
        \hline
    Rock 1& $14\degree$ & $0.194$ & $0.153$ & $0.187$ & $0.132$ & \textbf{0.109} & \textbf{0.081}\\
    $(5.7 \times 3.5 \times 2.8)$ &$28\degree$ &$0.202$& $0.159$ & $0.202$ & $0.159$ & \textbf{0.139} & \textbf{0.098} \\ 
    
         \hline
    Rock 2& $14\degree$ & $0.083$ & $ 0.065$ & $ 0.079 $ & $  0.060 $ & \textbf{0.071} & \textbf{0.056}\\
    $(2.2 \times 2.2 \times 2.0)$&$28\degree$ &$ 0.084 $& $ 0.065$ & $  0.082 $ & $ 0.063$ & \textbf{0.072} & \textbf{0.058} \\ 
    
         \hline
    
    Rock 3& $14\degree$ & $0.149$ & $0.093$ & $0.149$ & $0.098$ & \textbf{0.102} & \textbf{0.082}\\
    $(3.2 \times 3.7\times 2.8)$&$28\degree$ &$0.192$& $0.152$ & $0.166$ & $0.114$ & \textbf{0.148} & \textbf{0.103} \\ 
    
         \hline
         
    Coral& $14\degree$ & $0.241^*$ & $ 0.192^*$ & $0.241^*$ & $0.176^*$ & \textbf{0.134} & \textbf{0.106}\\
    $(4.4 \times 5.6 \times 3.3)$&$28\degree$ &$0.289^*$& $0.232^*$ & $0.285^*$ & $0.218^*$ & \textbf{0.212} & \textbf{0.166} \\     

         \hline
    Concrete column& $14\degree$ & $0.125$ & $0.097$ & $0.128$ & $0.099$ & \textbf{0.084} & \textbf{0.055}\\
    $(1.9 \times 1.2 \times 4.3)$&$28\degree$ &$0.149$& $0.113$ & $0.150$ & $0.115$ & \textbf{0.094} & \textbf{0.060} \\ 
    
             \hline
    Submarine& $14\degree$ & $0.187^*$ & $0.122^*$ & $0.204^*$ & $0.144^*$ & \textbf{0.173} & \textbf{0.101}\\
    $(5.1\times 16.7 \times 4.7)$ &$28\degree$ &$0.229^*$& $0.176^*$ & $0.237^*$ & $0.181^*$ & \textbf{0.149} & \textbf{0.102} \\

        \end{tabular}
    }
    \end{center}
    \caption{Size ($\text{W} \times \text{L} \times \text{H}$), root mean square (RMS) and mean Hausdorff distance errors (all in meters) for different simulated objects. For certain objects (*), we increased the voxel size from $0.025$m to $0.05$m to prevent OOM errors with the baseline methods.}
    \label{table:simtable}
\end{table}

\subsection{Water Tank Experiments}
\begin{figure}[h!]
\begin{subfigure}[h]{0.24\textwidth}
\includegraphics[width=0.8\columnwidth, height=3.5cm]{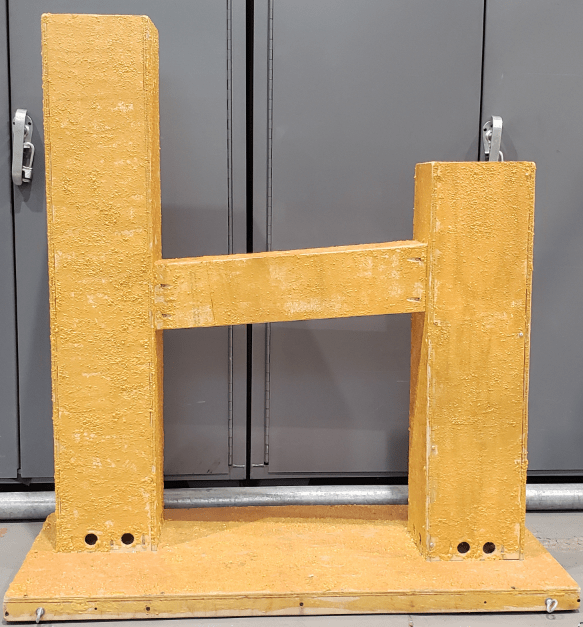}
\caption{Test structure}
\label{realstructure}
\end{subfigure}
\begin{subfigure}[h]{0.24\textwidth}
\includegraphics[width=\columnwidth, height=3.5cm]{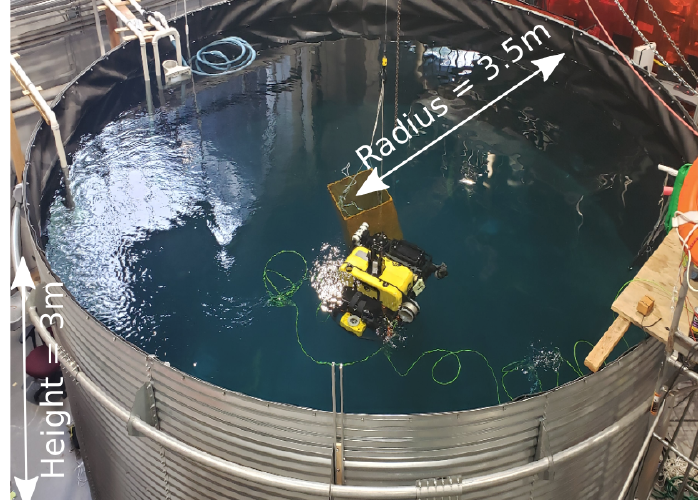}
\caption{Water test tank}
\label{testtank}
\end{subfigure}%
\caption{Real testing setup.}
\label{realexpsetup}
\vspace{-2mm}
\end{figure}

 \begin{figure} [h!]
 \centering
 \includegraphics[width=1\columnwidth]{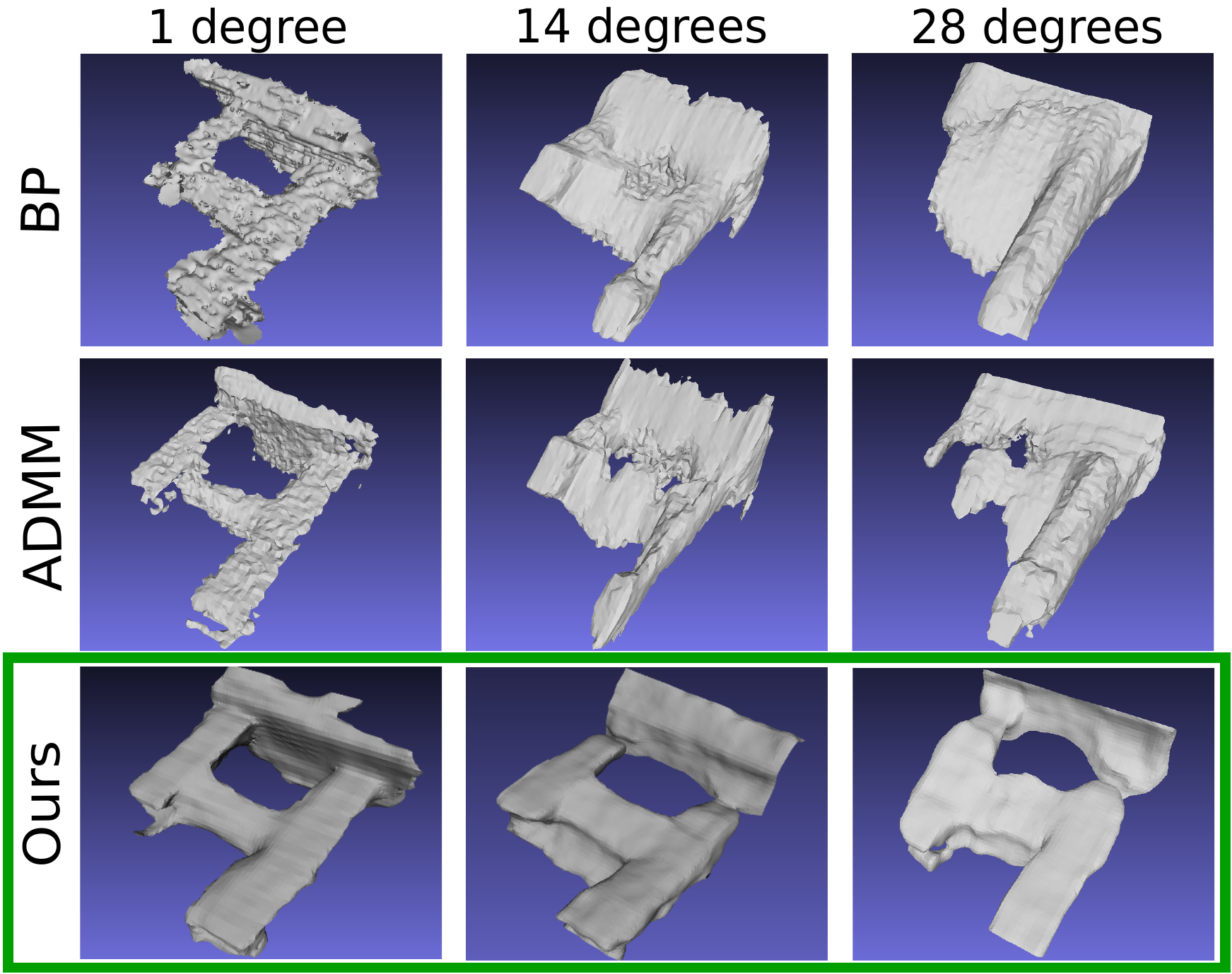}
\caption{The output 3D reconstructions for each method and each elevation aperture. Our method is able to capture the main components of the structures while ADMM and BP struggle for large vertical apertures.}
\label{fig:realQualitative}
\vspace{-4mm}
\end{figure}

\noindent
We evaluate our method on real-world datasets 
of a test structure (Fig.~\ref{realstructure}) submerged in a test tank (Fig.~\ref{testtank}) using a SoundMerics DIDSON imaging sonar mounted on a Bluefin Hovering Autonomous Underwater Vehicle (HAUV). The sonar can achieve three different elevation apertures ($1\degree, 14\degree, 28\degree$). We test our method on three different datasets, one for each feasible aperture. The vehicle uses a high-end IMU and a Doppler Velocity Log (DVL) to provide accurate vehicle pose information (i.e., minimal drift for the duration of data capture).

\begin{figure} [h!]
\centering
\includegraphics[width=1\columnwidth]{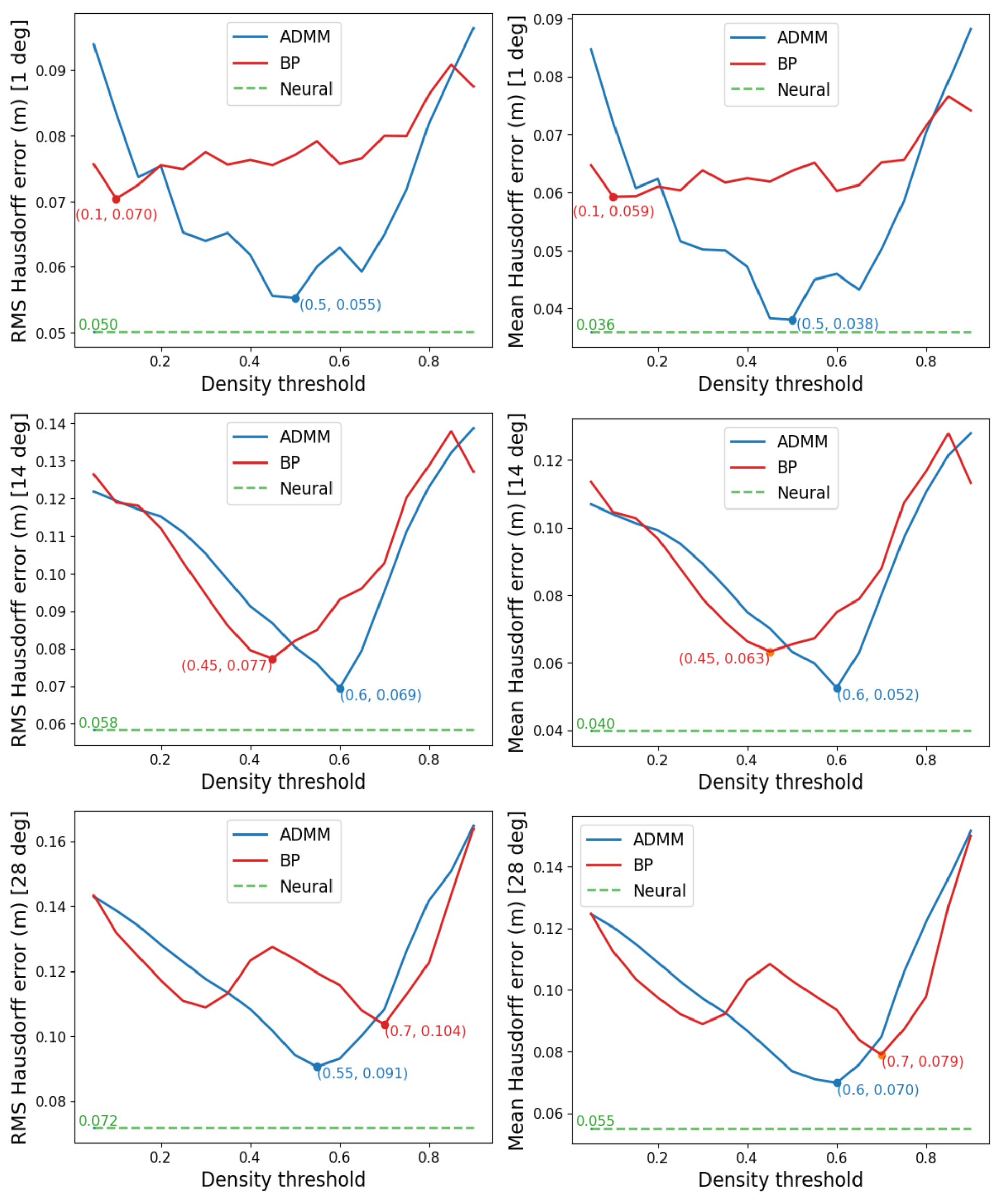}
\caption{Plots showing the root mean square (RMS) and mean Hausdorff distance in meters for all three methods on the real datasets ($1\degree, 14\degree, $ and $28\degree$ elevation apertures). To easily compare against the baselines, we add the constant green dashed line to report our method's metrics. Note however that our algorithm does not depend on the $\sigma$ values in the $x$ axis.}
\label{fig:realdataplots}

\end{figure}

Fig.~\ref{fig:realdataplots} shows the RMS and mean Hausdorff distance error of the three methods. The quality of the mesh generated by ADMM and BP depends on the selected marching cubes threshold $\sigma$. Hence, we plot the metrics generated using different $\sigma$s and report the best value. With our method, we can extract the zero-level set of $\mathbf{N}$ directly alleviating the need for a postprocessing step for surface generation. Since the structure is submerged and lying at the bottom of the test tank (and hence, no sonar image captures the backside of the object), we limit the matching distance of the Hausdorff metric to $0.15\text{m}, 0.2\text{m}, \text{ and } 0.25\text{m}$ for the $1\degree,14\degree, \text{ and } 28\degree$ apertures respectively. We see that our method generates higher quality reconstructions especially when using larger apertures: With $14\degree$, our method achieves an (RMS, Mean)=$(0.058\text{m}, 0.040\text{m})$ while BP and ADMM are respectively at $(0.077\text{m}, 0.063\text{m})$ and $(0.069\text{m}, 0.052\text{m})$. Similarly for a $28\degree$ aperture, our method achieves a lower (RMS, Mean) = $(0.072\text{m}, 0.055\text{m})$ compared to BP $(0.104\text{m}, 0.079\text{m})$ and ADMM $(0.091\text{m}, 0.070\text{m})$. 

Fig.~\ref{fig:realQualitative} shows the resulting meshes for each method. While all three methods perform well with a $1\degree$ aperture, the difference in reconstruction quality becomes visually more apparent as the aperture angle increases. With a $14 \degree$ aperture, we begin to lose the main feature of the object with ADMM and BP: the hole, short piling and crossbar are not easily discernible. When the aperture is increased to $28 \degree$, both baseline methods perform poorly: the hole, crossbar, and short piling are lost. On the other hand, our proposed method successfully captures the major components of the structure for all three different apertures (a base, two vertical pilings, and a crossbar).

\section{Conclusion and Future Work}
\noindent 
We proposed an approach for reconstructing 3D objects from imaging sonar which represents imaged surfaces as zero-level sets of neural networks. We performed experiments on simulated and real datasets with different elevation apertures and showed that our method outperforms current state-of-the-art techniques for unsupervised 3D reconstruction using FLS in terms of reconstruction accuracy. While existing volumetric methods can suffer from memory overhead as well as require a separate step to extract meshes from volumetric grids (a process often difficult and prone to error), our method allows for easy surface extraction from implicit representations and uses stochastic updates to lessen the computational requirements.

Our algorithm has some limitations, all of which create opportunities for future work. First, we currently focus on single-object reconstruction but plan to expand our method to large-scale reconstruction of marine environments at the scale of harbors by taking inspiration from techniques such as Block-Nerf \cite{tancik2022block}.  
Second, our method is currently mostly suited for offline 3D reconstructions but using techniques such as Instant-NGP  \cite{muller2022instant} and Relu-Fields \cite{karnewar2022relu} can bring it to real-time performance needed for robotic navigation applications.
Finally, all of our experiments now use only sonar but underwater robots are typically equipped with other sensors such as optical cameras. Hence, another interesting direction from future work is to fuse multi-modal sensor inputs (acoustic and optical) where, for example, optical cameras are used to obtain high resolution models of  specific interest areas in the scene while a sonar, with longer range, is used elsewhere.






\bibliographystyle{IEEEtran}
\IEEEtriggeratref{45}
\bibliography{main}

\end{document}